\documentclass[letterpaper,10pt,conference]{ieeeconf}
\IEEEoverridecommandlockouts
\usepackage{cite}
\usepackage{amsmath,amssymb,amsfonts}
\usepackage{graphicx}
\usepackage{booktabs}
\usepackage{array}
\usepackage{tabularx}
\usepackage{textcomp}
\usepackage{xcolor}
\usepackage{url}
\usepackage{etoolbox}
\usepackage{float}

\ifdefined
\fi

\floatstyle{ruled}
\newfloat{algorithm}{t}{loa}
\floatname{algorithm}{Algorithm}

\makeatletter
\apptocmd{\fst@algorithm}{\preto{\@fs@pre}{\vskip4pt}}
    {}{\PackageWarning{groove}{Could not inset the algorithm rule}}
\patchcmd{\@makecaption}
    {\footnotesize\scshape #2}
    {\footnotesize #2}
    {}{\PackageWarning{groove}{Could not patch the table-caption style}}
\makeatother

\graphicspath{{figures/}}

\newcommand{\method}{\textsc{Groove}}
\newcommand{\pifive}{$\pi_{0.5}$}
\newcommand{\grt}{GR00T N1.7}
\newcommand{\trans}{\mathrm{tr}}

\newcommand{\URRawSR}{46\%}
\newcommand{\URMethodSR}{52\%}
\newcommand{\URSR}{$+6.0$~pp}
\newcommand{\URSRCI}{$[-14.0,\,26.0]$~pp}
\newcommand{\URMcNemarP}{0.690}

\newcommand{\URTransTCPJerk}{16.39\%}
\newcommand{\URTransTCPJerkCI}{$[11.63,\,20.54]$\%}

\newcommand{\URRotTCPJerk}{19.49\%}
\newcommand{\URRotTCPJerkCI}{$[16.74,\,21.95]$\%}
\newcommand{\URPath}{6.89\%}
\newcommand{\URPathCI}{$[0.23,\,12.74]$\%}
\newcommand{\URCurrentSlew}{29.09\%}
\newcommand{\URCurrentSlewCI}{$[27.24,\,30.82]$\%}
\newcommand{\URFilterSensitivityTrans}{16.01--16.59\%}
\newcommand{\URFilterSensitivityRot}{17.89--21.88\%}
\newcommand{\URFilterSensitivityCurrent}{25.79--34.31\%}
\newcommand{\URJointAccelerationChange}{15.07\%}
\newcommand{\URJointAccelerationChangeCI}{$[9.59,\,19.53]$\%}

\newcommand{\URReplayReplans}{2,464}

\newcommand{\URReplayMean}{56.9~ms}

\newcommand{\URReplayPninetyFive}{69.0~ms}

\newcommand{\URReplayPninetyFivePeriodFraction}{24.2\%}

\newcommand{\URRawOnlinePostPolicyMean}{2.3~ms}
\newcommand{\UROnlinePostPolicyMean}{155.5~ms}
\newcommand{\UROnlinePostPolicyPninetyFive}{301.2~ms}

\newcommand{\URRawCommandIntervalMean}{289.9~ms}
\newcommand{\URMethodCommandIntervalMean}{310.3~ms}

\newcommand{\URCommandIntervalIncrease}{7.1\%}

\newcommand{\URStackSRChange}{$+12.0$~pp}

\newcommand{\URFullPaceTrans}{-1.74\%}
\newcommand{\URFullPaceTransCI}{$[-8.17,\,3.72]$\%}
\newcommand{\URFullPaceRot}{2.03\%}
\newcommand{\URFullPaceRotCI}{$[-2.11,\,5.85]$\%}

\newcommand{\URInteriorCommandIntervalIncrease}{0.5\%}

\newcommand{\URInteriorPaceOnly}{1.17\%}

\newcommand{\URInteriorTrans}{14.97\%}
\newcommand{\URInteriorTransCI}{$[9.49,\,19.84]$\%}
\newcommand{\URInteriorRot}{18.65\%}
\newcommand{\URInteriorRotCI}{$[15.39,\,21.83]$\%}

\newcommand{\URInteriorCurrentSlew}{29.44\%}
\newcommand{\URInteriorCurrentSlewCI}{$[27.51,\,31.24]$\%}
\newcommand{\URInteriorCurrentSlewPaceOnly}{0.39\%}

\newcommand{\PiCompFixedGrooveTrans}{39.19\%}
\newcommand{\PiCompFixedGrooveTransCI}{$[35.96,\,42.69]$\%}
\newcommand{\PiCompFixedGrooveRot}{49.08\%}
\newcommand{\PiCompFixedGrooveRotCI}{$[46.29,\,51.84]$\%}

\newcommand{\PiTransJerk}{32.45\%}

\newcommand{\PiRotJerk}{43.91\%}

\newcommand{\PiSR}{$+0.25$\,pp}

\newcommand{\GRTransJerk}{37.02\%}

\newcommand{\GRRotJerk}{53.42\%}

\newcommand{\GRSR}{$+0.75$\,pp}

\newcommand{\PiTraceArtifactDigest}{UNSET}
\newcommand{\PiTraceTableRows}{UNSET}
\newcommand{\PiTraceHeadlineSentence}{UNSET}
\newcommand{\PiTracePrimarySentence}{UNSET}
\newcommand{\PiTraceSeamSentence}{UNSET}
\newcommand{\PiTraceComparatorSentence}{UNSET}
\newcommand{\PiTraceMechanismSentence}{UNSET}
\newcommand{\PiTraceFidelitySentence}{UNSET}
\newcommand{\PiTraceRuntimeSentence}{UNSET}
\newcommand{\PiTraceConclusionSentence}{UNSET}
\newcommand{\PiTraceRuntimeCalls}{UNSET}
\newcommand{\PiTraceLatencyPninetyNine}{UNSET}
\newcommand{\PiTraceDeadlineMissCount}{UNSET}
\renewcommand{\PiTraceArtifactDigest}{0626f0c69af29a98faa78165f7dc5c559cd1d26fe7e357e3d0473264a918ba00}
\renewcommand{\PiTraceTableRows}{Single Tube vs. raw & $+38.52\%~[35.58,\,41.59]$ \\
Single Tube vs. EMA & $+16.14\%~[12.93,\,19.37]$ \\
Single Tube vs. One Euro & $+16.53\%~[13.40,\,19.70]$ \\
History vs. no history$^\dagger$ & $+35.20\%~[31.53,\,39.05]$ \\
Prefix vs. per-command guard & $+10.28\%~[8.29,\,12.25]$ \\
Prefix vs. endpoint guard$^\dagger$ & $-33.59\%~[-38.80,\,-28.51]$ \\}
\renewcommand{\PiTraceHeadlineSentence}{A separate 400-pair \pifive{} single-tube holdout reduces fixed-50 translational EEF jerk by $+38.52\%~[35.58,\,41.59]$ and changes paired success by $+3.25~[0.50,\,6.75]$~pp. It also outperforms budget-matched EMA, One Euro, and the per-command guard.}
\renewcommand{\PiTracePrimarySentence}{In the single-tube holdout, fixed-50 translational EEF jerk shows a $+38.52\%$ reduction, and paired success changes by $+3.25$~pp.}
\renewcommand{\PiTraceSeamSentence}{Seam and interior reductions are $+42.40\%$ and $+36.56\%$, respectively. Their interaction is $0.097$, indicating a larger proportional reduction at seams.}
\renewcommand{\PiTraceComparatorSentence}{Against budget-calibrated filters, fixed-50 reductions are $+16.14\%~[12.93,\,19.37]$ versus EMA and $+16.53\%~[13.40,\,19.70]$ versus One Euro. Both comparisons have positive 95\% CI lower bounds.}
\renewcommand{\PiTraceMechanismSentence}{Against the budget-matched per-command guard, the prefix-wide guard reduces fixed-50 jerk by $+10.28\%~[8.29,\,12.25]$ with a paired success change of $+3.25~[0.00,\,7.00]$~pp. Its seam-localization interaction CI [-0.087,\,-0.005] places the advantage across the full window rather than specifically at seams. Removing history yields a descriptive seam-localized effect of $0.487~[0.433,\,0.542]$. Endpoint-only records 25.14\% lower jerk than single-tube with 90.7\% greater realized deviation budget and 4.75~pp lower success. This unmatched row characterizes a jerk--fidelity tradeoff rather than superiority.}
\renewcommand{\PiTraceFidelitySentence}{Across 12,973 single-tube replans in the 400-pair holdout, the maximum logged controller-command prefix deviation is 0.15000011 at $\epsilon=0.15$: an excess of $1.1\!\times\!10^{-7}$, below the $10^{-5}$ audit tolerance, with zero solver fallbacks.}
\renewcommand{\PiTraceRuntimeCalls}{12,973}
\renewcommand{\PiTraceLatencyPninetyNine}{14.87}
\renewcommand{\PiTraceDeadlineMissCount}{0}
\renewcommand{\PiTraceRuntimeSentence}{Across \PiTraceRuntimeCalls{} single-tube calls, p99 latency is \PiTraceLatencyPninetyNine{}~ms with \PiTraceDeadlineMissCount{} 50-ms deadline misses. Timing excludes VLA inference and environment execution.}
\renewcommand{\PiTraceConclusionSentence}{The fixed-duration holdout confirms the effect against calibrated filters and the matched per-command guard.}

\newcommand{\RequirePiTraceSlot}[2]{%
    \ifdefstring{#1}{UNSET}{%
        \PackageError{GROOVE}{Evidence slot #2 is unset}%
        {Render #2 from the integrity-complete pi0.5 trace report.}%
    }{}}

\AtBeginDocument{%
    \RequirePiTraceSlot{\PiTraceArtifactDigest}{PiTraceArtifactDigest}%
    \RequirePiTraceSlot{\PiTraceTableRows}{PiTraceTableRows}%
    \RequirePiTraceSlot{\PiTraceHeadlineSentence}{PiTraceHeadlineSentence}%
    \RequirePiTraceSlot{\PiTracePrimarySentence}{PiTracePrimarySentence}%
    \RequirePiTraceSlot{\PiTraceSeamSentence}{PiTraceSeamSentence}%
    \RequirePiTraceSlot{\PiTraceComparatorSentence}{PiTraceComparatorSentence}%
    \RequirePiTraceSlot{\PiTraceMechanismSentence}{PiTraceMechanismSentence}%
    \RequirePiTraceSlot{\PiTraceFidelitySentence}{PiTraceFidelitySentence}%
    \RequirePiTraceSlot{\PiTraceRuntimeSentence}{PiTraceRuntimeSentence}%
    \RequirePiTraceSlot{\PiTraceConclusionSentence}{PiTraceConclusionSentence}%
    \RequirePiTraceSlot{\PiTraceRuntimeCalls}{PiTraceRuntimeCalls}%
    \RequirePiTraceSlot{\PiTraceLatencyPninetyNine}{PiTraceLatencyPninetyNine}%
    \RequirePiTraceSlot{\PiTraceDeadlineMissCount}{PiTraceDeadlineMissCount}%
}

\begin{document}

\bstctlcite{IEEEexample:BSTcontrol}

\title{\LARGE\bfseries GROOVE: Geometry-Guided Reduction of\\
Operational-Space Jerk in VLA Execution}

\urldef{\grooveProjectURL}\url{https://devsangho.github.io/GROOVE-public/}
\author{\authorblockN{Sangho Yun$^{1}$, Minsoo Kim$^{2}$, Minwoo Cho$^{1}$, and Hwanjo Yu$^{1,2}$}%
\thanks{$^{1}$Department of Computer Science and Engineering,
POSTECH, Pohang, South Korea.}%
\thanks{$^{2}$Graduate School of Artificial Intelligence,
POSTECH, Pohang, South Korea.}%
\thanks{Correspondence to: Hwanjo Yu
\mbox{\textless hwanjoyu@postech.ac.kr\textgreater}.}%
\thanks{Project page: \grooveProjectURL.}}

\AddToHookNext{shipout/foreground}{%
    \put(0,-36){\makebox[\paperwidth][c]{%
        \parbox{\textwidth}{\centering\normalfont\footnotesize
        This work has been submitted to the IEEE for possible publication.\\
        Copyright may be transferred without notice, after which this version may no longer be accessible.}%
    }}%
}

\maketitle

\begin{abstract}
Chunked vision--language--action (VLA) policies execute several commands per query, but jerk within chunks and across replanning boundaries can induce oscillatory motion and sharp actuator transients.
We present \method{}, an online regulator that searches directional correction regions around the raw three-dimensional end-effector (EEF) path, without retraining or additional VLA inference.
It optimizes the new chunk using delivered commands as boundary conditions, reducing boundary and within-chunk jerk while bounding cumulative translation and local axis--angle deviation from the raw plan after every command.
Using quadratic programs (QPs), \method{} generates a cube reference and thirteen directional candidates, then selects the one with the lowest command-space jerk under a reference-relative deviation cap.
On a held-out LIBERO benchmark, \method{} achieves the largest reductions among the evaluated methods, reducing translational and rotational EEF jerk by 33.02\% and 43.42\%, respectively, with task success of 95.75\% versus 93.75\% for raw execution.
Across 50 matched UR5e pairs with measured execution timing, it reduces translational and rotational tool-center-point (TCP) jerk by 16.39\% and 19.49\% and joint-current slew by 29.09\%.
\end{abstract}


\section{Introduction}

VLA policies execute single actions or multi-step chunks~\cite{openvla,act,pi05,gr00t}.
Chunking lets one query cover several control steps while promoting temporal coherence~\cite{act,pi05,gr00t,a2c2}.
However, chunked execution can introduce jerk through two distinct mechanisms.
Within a chunk, oscillations in predicted delta end-effector (EEF) commands can cause rapid motion changes.
At a replanning boundary, the initial commands of a new chunk may not align with those most recently delivered to the controller~\cite{seam,rtc,potr,lipo}.
Aligning only the replanning boundary therefore leaves within-chunk jerk unaddressed.
Task success alone does not distinguish smooth execution from execution with abrupt motion and actuator transients.
On physical manipulators, abrupt acceleration changes can excite structural vibration, degrade path tracking, and produce sharp torque and power transients~\cite{bearee2014dampedjerk,lee2024jerkconstrained}.
For a policy that already completes a task, reducing these transients is an additional objective that requires limiting changes to task-directed motion and accounting for execution time.
We therefore evaluate measured motion and joint-current variation alongside task success and timing~\cite{embodiedefficiency}.

Existing approaches improve temporal consistency by modifying policy training or action generation~\cite{smoothvla,rtc,seam,potr}.
Training-time methods require retraining, while generation-time methods can require access to policy internals or new predictions during execution.
Overlap-based methods depend on multiple unexecuted predictions, and ACT creates this overlap by querying the VLA at every step~\cite{act,lipo}.
Post-generation filters avoid these requirements by modifying commands after generation~\cite{oneeuro}.
However, local corrections to delta-EEF commands accumulate as the commands are integrated into an EEF path.
A filter can therefore reduce jerk while allowing the corrected path to drift from the raw task-directed path.
Constraining only the chunk endpoint does not solve this problem because intermediate path deviations remain unconstrained.

\begin{figure}[!t]
    \centering
    \includegraphics[width=\columnwidth]{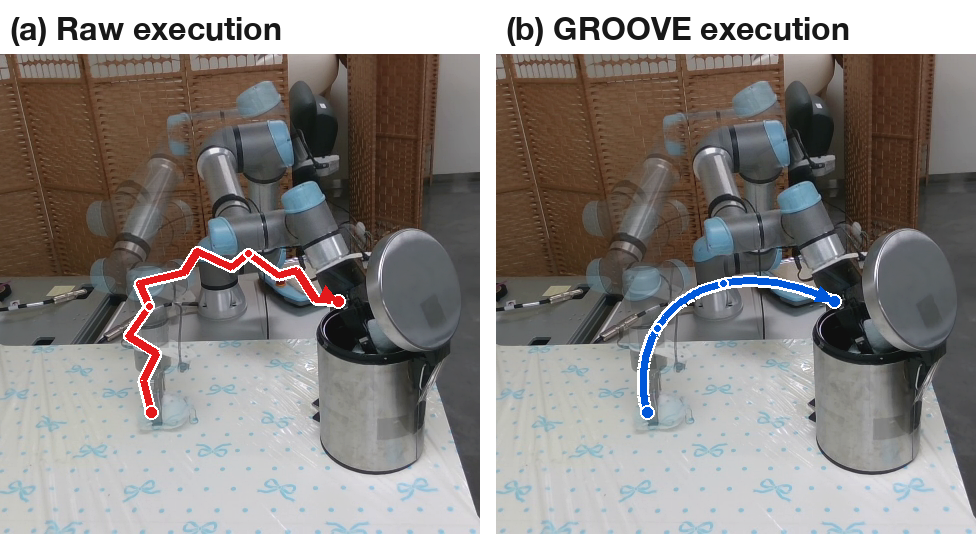}
    \caption{Conceptual comparison of raw and \method{} execution.}
    \label{fig:replan-teaser}
\end{figure}

\begin{figure*}[!t]
    \centering
    \includegraphics[width=\textwidth]{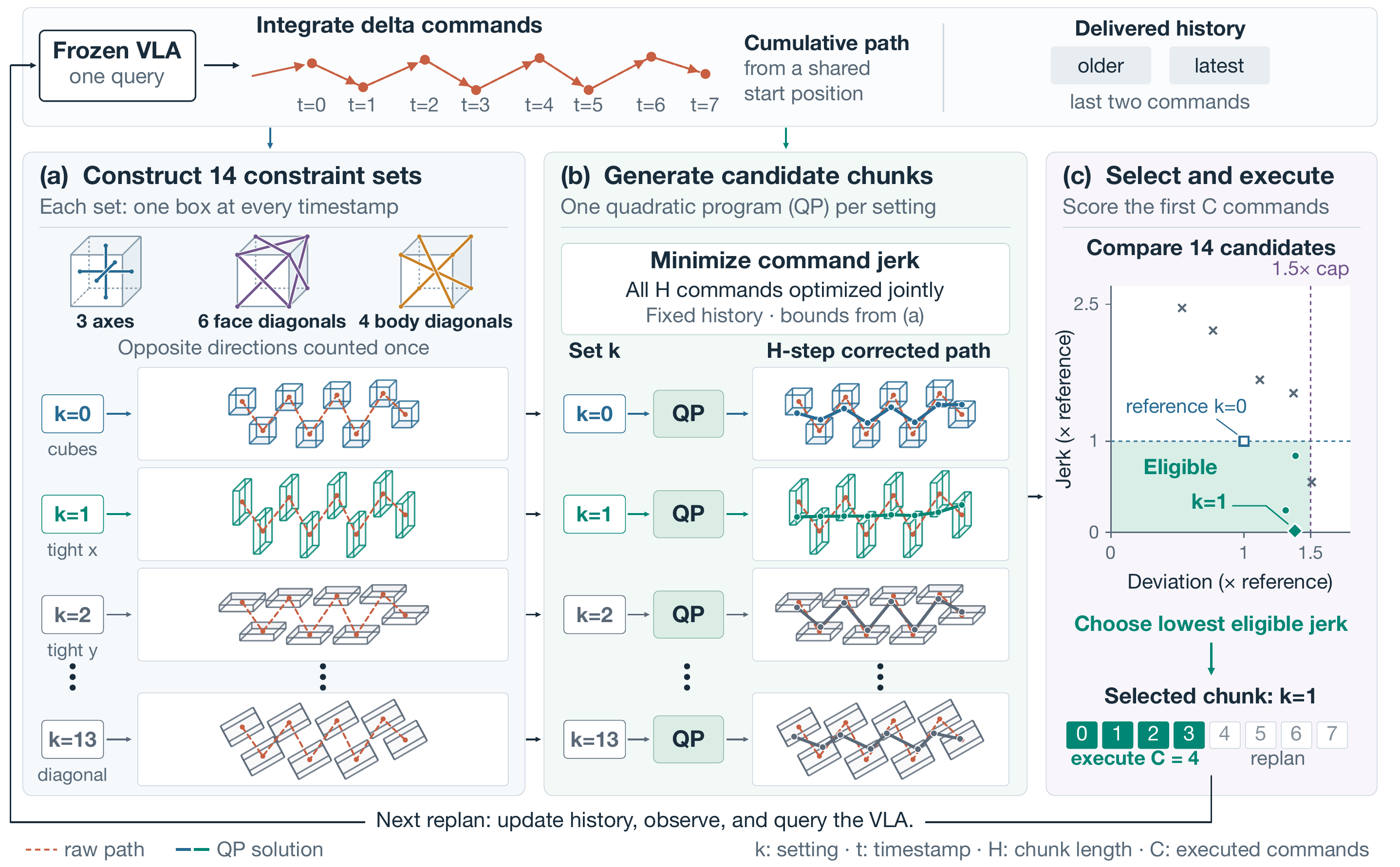}
    \caption{\method{} at one replan in a synthetic example computed by the QPs ($H=8$, $C=4$). (a) Construct 14 constraint sets, each placing a box at every raw-path timestamp. (b) Generate candidate chunks by jointly optimizing all $H$ commands within the corresponding boxes, using delivered commands as boundary conditions. Corrected paths are overlaid on the red raw paths. (c) Select and execute using deviation and jerk over the first $C$ commands. The blue square denotes the retained reference, crosses mark ineligible candidates, and the teal diamond marks the selected candidate $k=1$. All fourteen score pairs are normalized by the reference, and some overlap.}
    \label{fig:method-overview}
\end{figure*}

We introduce \method{} (Geometry-Guided Reduction of Operational-Space Jerk in VLA Execution), which regulates a new action chunk by searching the geometry of allowable corrections around its raw three-dimensional EEF path.
A cube applies the same bound along each coordinate axis even when corrections that reduce jerk are concentrated along particular directions or planes.
\method{} constructs a cube reference and thirteen equal-volume directional constraint sets, each defining a different correction region at every raw-path timestamp (Figs.~\ref{fig:replan-teaser} and~\ref{fig:method-overview}).
A full-chunk quadratic program (QP) computes a candidate for each set using the last two delivered commands as boundary conditions.
The selector chooses the candidate with the lowest command-space jerk under a reference-relative deviation cap.
This geometric search runs after one VLA query, with every-prefix translation and local-coordinate rotation bounds and no retraining or additional prediction.

We make three contributions.
\begin{itemize}
    \item We propose a geometric search over directional correction regions around the raw three-dimensional EEF path, with protected candidate selection under a reference-relative deviation cap.
    \item We implement this search through full-chunk optimization conditioned on delivered commands, with translation and rotation bounds at every prefix and no additional VLA inference.
    \item We evaluate the method with two frozen VLAs and on a UR5e. It yields the largest holdout jerk reductions among the evaluated methods and lowers TCP jerk and current slew on the robot, with higher aggregate success point estimates.
\end{itemize}

\section{Related Work}

\subsection{Jerk-Aware Trajectory Generation}

Minimum-jerk models optimize an entire trajectory by minimizing integrated squared jerk~\cite{flashhogan}.
Jerk-limited motion is also used to suppress structural vibration, improve tracking, and reduce peak actuator demands on physical manipulators~\cite{bearee2014dampedjerk,lee2024jerkconstrained}.
Ruckig computes time-optimal motion under velocity, acceleration, and jerk limits~\cite{ruckig}.
These methods generate motion from boundary and target states.
Regulating intermediate VLA delta-action paths requires additional deviation constraints.

\subsection{Chunked Action Prediction and Aggregation}

Action chunking reduces the average inference cost per control step by predicting several future commands in one query.
OpenVLA predicts one action at a time, whereas \pifive{} and GR00T produce action chunks~\cite{openvla,pi05,gr00t}.
ACT predicts action chunks and temporally aggregates overlapping predictions at each control step~\cite{act}.
BID samples multiple candidate chunks at test time and ranks them by consistency with past actions and contrast among future candidates~\cite{bid}.
ACT and BID combine multiple predictions to improve temporal consistency, but neither directly minimizes jerk or bounds cumulative EEF deviation from the raw path.

\subsection{Training-Time Smoothness}

SmoothVLA modifies a policy through reinforcement learning with a reward that combines task success and trajectory jerk~\cite{smoothvla}.
Legato trains a flow policy to connect each new chunk smoothly to preceding actions~\cite{legato}.
RTR learns a VAE-based latent action space for high-frequency chunks and refines concatenated executed and predicted actions through the VAE at inference time~\cite{rtr}.
ChunkFlow uses seam and first- and second-order continuity losses during training and deterministic overlap blending during execution~\cite{chunkflow}.
These approaches improve continuity by retraining the policy or learning an additional action representation.

\subsection{Sampling-Time Action Consistency}

Sampling-time methods modify how the policy samples actions to improve continuity across chunks.
RTC uses flow inpainting to align the prefix of a new chunk with the unexecuted suffix of the previous chunk~\cite{rtc}.
PAINT selects initial noise that improves prefix consistency before denoising~\cite{paint}.
SEAM replaces RTC's backward passes with a closed-form correction toward the previous suffix~\cite{seam}.
POTR modifies RTC guidance using prior correction and an orthogonal trust region~\cite{potr}.
ACG applies action-coherence guidance at selected attention layers during flow sampling~\cite{acg}.
These methods avoid retraining but require access to policy sampling.
RTC, SEAM, and POTR require the previous unexecuted suffix, which is unavailable after full-chunk execution and contains planned rather than delivered commands.
ACG instead evaluates an extra incoherent vector field during sampling.

\subsection{Post-Generation Regulation}

Post-generation methods regulate VLA outputs without retraining the policy.
EMA uses a fixed decay, while One Euro adapts its low-pass cutoff to signal speed~\cite{oneeuro}.
PACE sets execution length from low-speed transitions in the predicted chunk~\cite{pace}.
AAC shortens chunks under high action entropy and lengthens them under low entropy~\cite{aac}.
A2C2 applies a trained time-aware correction head at each control step~\cite{a2c2}.
LiPo minimizes jerk within per-step joint-position bounds around blended, overlapping trajectories~\cite{lipo}.
Realtime-VLA V2 combines adaptive timing with model-based joint tracking~\cite{realtimevlav2}.
\method{} searches three-dimensional correction regions around the raw EEF path.
A cube reference and equal-volume directional boxes define candidate paths, which are optimized from one new chunk and delivered-command history.
A reference-relative selector chooses among these corrections using their jerk and path deviation.

\section{Method}
\label{sec:groove-method}

\method{} reduces jerk by searching directional correction regions around the raw three-dimensional EEF path.
Figure~\ref{fig:method-overview} shows how we define allowed corrections (a), optimize a candidate chunk under each setting (b), and select a candidate for execution (c).
Each candidate contains $H$ commands, of which the first $C$ are executed before replanning ($1\leq C\leq H$).

\subsection{Construct 14 Constraint Sets}
\label{sec:construct-tubes}

Smoothing individual delta commands can shift the path obtained by accumulating them.
To limit this drift throughout the chunk, we place a box around each raw cumulative position and require the corrected position at that step to remain inside its corresponding box.
Each row in Fig.~\ref{fig:method-overview}(a) defines one constraint set containing $H$ boxes.
This sequence of boxes forms a tube around the raw path.
The arrows represent commands, and the points represent cumulative positions.

Corrections that reduce jerk can be larger in some directions than others.
We therefore compare a cube reference ($k=0$) with thirteen settings that allow corrections along different directions ($k=1,\ldots,13$).
Each directional box is narrower along one axis and wider along the two perpendicular axes, with the same volume as the reference.
For example, the $k=1$ row restricts $x$ deviation while allowing more correction along $y$ and $z$.
The fixed set of thirteen tight axes covers three coordinate axes, six face diagonals, and four body diagonals.
It is independent of the raw chunk, which determines only the box centers.
The box axes remain fixed throughout each row.

Near gripper transitions, all settings use smaller boxes to limit changes around opening or closing commands.
The gripper commands themselves remain unchanged.

\subsection{Generate Candidate Chunks}
\label{sec:full-chunk-optimization}

To reduce abrupt changes both within a chunk and at its start, we optimize all commands together using the last two delivered commands as boundary conditions.
Only the box constraints vary across settings.

Let $r$ be the raw arm chunk and $v$ the chunk to optimize.
At step $t$, each command $r_t,v_t\in\mathbb R^6$ contains controller-normalized translation and local axis--angle increments, denoted by superscripts $\trans$ and $\mathrm{rot}$.
The delivered commands $h_{-2},h_{-1}$ fix $v_{-2}=h_{-2}$ and $v_{-1}=h_{-1}$, with missing history initialized to zero.
We penalize the second difference $\delta_h^2v_t=v_t-2v_{t-1}+v_{t-2}$.
At $t=0$, the second difference $v_0-2h_{-1}+h_{-2}$ connects the new chunk to delivered commands.
Because translation commands are position increments, their second differences are proportional to position jerk under ideal integration at a fixed command period.

To express the boxes in (a), define the signed cumulative command difference $e_t(v)=\sum_{i=0}^{t}(v_i-r_i)$.
Its translation component points from the raw cumulative position to the corrected position.
For setting $k$, the columns of the orthonormal matrix $Q_k$ are the three box axes, and $\eta_{k,t}$ contains their half-widths, measured from center to face.
Thus $Q_k^\top e_t^{\trans}(v)$ expresses the deviation along those axes.
Rotation uses a shared half-width $\epsilon_t$, and $\mathbf1_3$ denotes the three-component all-ones vector.
With a fixed weight $\lambda\geq0$ on command magnitude, the QP returns candidate $u_k$,
\begin{equation}
\begin{aligned}
    u_k&=\arg\min_{v_{0:H-1}}\sum_{t=0}^{H-1}\left(
        \left\lVert\delta_h^2v_t\right\rVert_2^2
        +\lambda\lVert v_t\rVert_2^2\right),\\
    &\text{s.t. }\left|Q_k^\top e_t^{\trans}(v)\right|\preceq\eta_{k,t},\\
    &\phantom{\text{s.t. }}\left|e_t^{\mathrm{rot}}(v)\right|\preceq\epsilon_t\mathbf{1}_3,
        \quad t=0,\ldots,H-1.
\end{aligned}
    \label{eq:full-chunk-qp}
\end{equation}
Absolute values and inequalities apply componentwise.
Second differences and cumulative constraints couple the $H$ steps, so all $6H$ components are optimized jointly.
Solving this QP for each row produces the fourteen candidate paths in (b).

\subsection{Select and Execute}
\label{sec:select-execute}

A candidate can reduce jerk while increasing deviation, even when its boxes have the same volume as the reference.
We therefore limit deviation relative to the reference and select the smoothest candidate that meets this limit.
Only the first $C$ commands enter this comparison because they will be executed before replanning.
Rotation is shared across candidates, so we compare translation.
For candidate $u_k$, with command $u_{k,t}$ at step $t$, define
\begin{equation}
\begin{aligned}
    B_k&=\sum_{t=0}^{C-1}\left\lVert e_t^{\trans}(u_k)\right\rVert_1,\\
    P_k&=\frac{1}{C}\sum_{t=0}^{C-1}
        \left\lVert\delta_h^2u_{k,t}^{\trans}\right\rVert_2.
\end{aligned}
    \label{eq:selection-scores}
\end{equation}
The deviation score $B_k$ sums the L1 norms of cumulative translation deviations in the original coordinates, capturing intermediate drift even if later corrections cancel it.
The jerk score $P_k$ averages translation second-difference norms over the executed prefix, whereas the QP sums squared norms over all $H$ arm commands.

The cube setting yields the QP-generated reference $u_0$, with scores $B_0,P_0$.
If the reference fails numerical validation, we use the raw chunk.
Otherwise, we retain it and admit a directional candidate only if it is numerically valid, satisfies $B_k\leq\rho_B\max(B_0,10^{-12})$, and has $P_k<P_0$.
The fixed, dimensionless multiplier $\rho_B\geq1$ sets the allowed reference-relative deviation, with $10^{-12}$ serving as a numerical floor.
We choose the candidate with the smallest $P_k$ among the reference and admitted candidates, using a fixed ordering for exact ties.
This ensures that the selected candidate has a jerk score of at most $P_0$ and satisfies the deviation cap.
These guarantees concern command-space scores, with physical motion evaluated separately.
Panel~(c) plots $(B_k/B_0,P_k/P_0)$ with positive reference denominators and selects $k=1$ when $\rho_B=1.5$.

The execution length $C$ is shared with raw execution and all comparators for each policy.
Regulation does not shorten $C$ or query the VLA again.
We send the first $C$ arm commands from the chosen chunk with the corresponding original gripper commands $g_t$ and discard the remaining $H-C$ commands.
At the next replan, we update history from acknowledged commands and query the VLA using the new observation.
Algorithm~\ref{alg:groove} summarizes the procedure.

\subsection*{Implementation Details}

Outside gripper transitions, the reference uses the identity frame $Q_0=I_3$ and half-widths $\eta_{0,t}=\epsilon_0\mathbf1_3$.
Directional settings use $\eta_{k,t}=[\epsilon_{\mathrm{tight}},\epsilon_{\mathrm{wide}},\epsilon_{\mathrm{wide}}]^\top$, with $0<\epsilon_{\mathrm{tight}}<\epsilon_0<\epsilon_{\mathrm{wide}}$ and $\epsilon_{\mathrm{tight}}\epsilon_{\mathrm{wide}}^2=\epsilon_0^3$ to preserve box volume.
Their tight axes are the normalized nonzero vectors in $\{-1,0,1\}^3$, counted once per opposite pair, and form the first columns of $Q_k$.
Choose the coordinate axis least aligned with the first column, breaking ties in $x,y,z$ order.
Project this axis onto the plane perpendicular to the first column and normalize the projection to obtain the second column.
The third column is the cross product of the first two.

Within $w_g$ steps of a sign change in $g_t$, including transitions from the last delivered gripper command, each setting uses $\eta_{k,t}=\epsilon_g\mathbf1_3$ in its own frame.
The shared rotation half-width $\epsilon_t$ equals $\epsilon_g$ in this window and $\epsilon_0$ elsewhere.
It bounds cumulative local axis--angle commands rather than exact orientation error.

In the orthonormal box frame, the objective is unchanged and translation separates into three scalar QPs spanning $H$ steps.
Reusing the common rotation solution gives $14\times3+3=45$ scalar solves.
Fixed history makes the objective strictly convex even at $\lambda=0$, and the raw chunk is feasible, so each QP has a unique exact solution.
Numerical validation checks every prefix bound and exact preservation of gripper commands.

\begin{algorithm}[!t]
\caption{\method{} at one replan.}
\label{alg:groove}
\footnotesize
\begin{tabular}{@{}r@{\hspace{0.45em}}p{0.88\columnwidth}@{}}
\multicolumn{2}{@{}l}{\textbf{Input:} raw chunk, delivered arm history, and last gripper state.}\\
1 & \textbf{A. Construct 14 constraint sets.} Prepare the boxes in Fig.~\ref{fig:method-overview}(a) and shared rotation bounds.\\
2 & \textbf{B. Generate candidate chunks. For} $k=0,\ldots,13$ \textbf{do}\\
3 & \quad Solve Eq.~\eqref{eq:full-chunk-qp} for the entire chunk to obtain $u_k$.\\
4 & \quad Validate every prefix bound and unchanged gripper commands.\\
5 & \textbf{end for}\\
6 & \textbf{C. Select and execute.} If the reference is invalid, use the raw chunk.\\
7 & Otherwise, compute $B_k,P_k$ on the first $C$ commands and admit valid candidates meeting the conditions in Sec.~\ref{sec:select-execute}.\\
8 & \quad Choose the candidate with the smallest $P_k$ among the reference and admitted candidates.\\
9 & Send the first $C$ commands of the chosen chunk with the original gripper commands.\\
10 & Update delivered history and replan from the next observation.\\
\end{tabular}
\end{algorithm}

\section{Experiments}

We compare jerk and task success under matched query budgets, evaluate physical effects and timing, and examine ablations and runtime.

\subsection{Simulation Benchmark}

We evaluate frozen \pifive{} and \grt{} policies on all 40 tasks from LIBERO-Spatial, LIBERO-Object, LIBERO-Goal, and LIBERO-10 at 20~Hz~\cite{libero,pi05,gr00tn17libero}.
We use prediction horizons and execution lengths of $(H,C)=(10,5)$ for the OpenPI pi05\_libero checkpoint and $(16,8)$ for \grt{}.
Development grids contain 800 matched raw--\method{} pairs per policy for comparisons across policies under shared settings and for mechanism analyses.
The primary comparison uses a separately frozen \pifive{} holdout of 400 matched episodes, with 100 per suite.

Within each matched block, methods share the policy checkpoint, task, initial state, sampling seed, and episode command limit.
All holdout methods, including raw execution, use $(H,C)=(10,5)$ and one VLA query per five commands, with no extra observations or predictions. Command limits are 220, 280, 300, and 520 for Spatial, Object, Goal, and LIBERO-10, respectively.
Methods with equal query budgets can have different inference and post-processing latencies, so we assess physical timing separately.
We retain every assigned episode, and holdout outcomes were unavailable during method selection.

\subsection{Comparators and Ablation Controls}

The main holdout compares raw execution with RTC, a SEAM sweep over $\lambda\in\{0.05,0.10,0.20\}$, POTR, a query-matched overlap ensemble, a LiPo-QP adapter evaluated under the same protocol, and \method{}~\cite{rtc,seam,potr,act,lipo}.
RTC follows the authors' $\Pi$GDM prior-guidance formulation with a five-step prefix and maximum guidance of 5.
SEAM guides all five available overlap steps, with $M=\min(20,H-C)=5$ at each setting. POTR uses the paper's LIBERO values $\sigma_d=0.4$, $\rho=0.5$, and $\beta=10$.
Each configuration was frozen before evaluation.
These generation-time methods leave the first chunk unguided because no previous suffix is available.
We implement POTR in OpenPI following Eqs.~5--7 of the original paper~\cite{potr}.

The overlap ensemble averages the previous raw suffix and aligned new prefix with weights $\exp(-0.01i)$, where $i=0$ denotes the previous suffix and $i=1$ the new prefix~\cite{act}.
The LiPo-QP adapter integrates delta-EEF commands, applies LiPo's overlap blend and ActionLiPo QP, and converts the result back to delta commands~\cite{lipo}.
Its frozen overlap and path bounds are $0.4$ and $0.06$ in cumulative controller-normalized coordinates, preserving the original $0.02/0.003$ ratio at a scale of 20.
Both adapters retain the same query rate, controller interface, and command budget. The LiPo adapter omits its controller-specific 400-Hz interpolation.

A follow-up candidate-set ablation reruns raw execution and four sets on the 400 initial-state assignments, using a separately frozen inference runtime and the same suite command limits.
The sets comprise the cube alone, cube plus three axes, cube plus four body diagonals, and all fourteen candidates, with fixed QP parameters, selector, and $(H,C)=(10,5)$.
A separate 400-triplet rollout compares the cube with full sets at $\rho_B=1$ and $1.5$, with otherwise fixed settings.
A post-hoc replay adjusts cube translation width to match selected first-$C$ deviation $B$ on stored full-set inputs and delivered history, retaining the objective and gripper guard.

On the development grid, we compare the full set with the cube reference, one direction sampled at each replan, and one candidate sampled uniformly from the eligible set.
Both sampling controls use an episode-seeded generator.
A sensitivity analysis restricted to the first 50 commands fixes the trajectory length.
Fixed-window controls compare delivered-history, per-command, and endpoint-only variants with budget-matched EMA and One Euro filters~\cite{oneeuro}.
Filter strengths are frozen on a disjoint 40-pair calibration set by matching the cube regulator's mean realized prefix deviation per replan without using success or jerk.
Across all policies, \method{} uses $\epsilon_0=0.15$, $\epsilon_{\mathrm{tight}}=0.05$, $\epsilon_{\mathrm{wide}}=\sqrt{\epsilon_0^3/\epsilon_{\mathrm{tight}}}\approx0.2598$, $\epsilon_g=0.001$, $w_g=5$, $\lambda=5\times10^{-9}$, and $\rho_B=1.5$ in the controller-normalized coordinates defined above.
\method{} uses OSQP with absolute and relative tolerances of $10^{-6}$, a 4,000-iteration limit, and a feasibility tolerance of $10^{-5}$~\cite{osqp}.

\subsection{Physical-Robot Protocol}

\begin{figure}[t]
    \centering
    \vspace*{4pt}
    \includegraphics[width=\columnwidth]{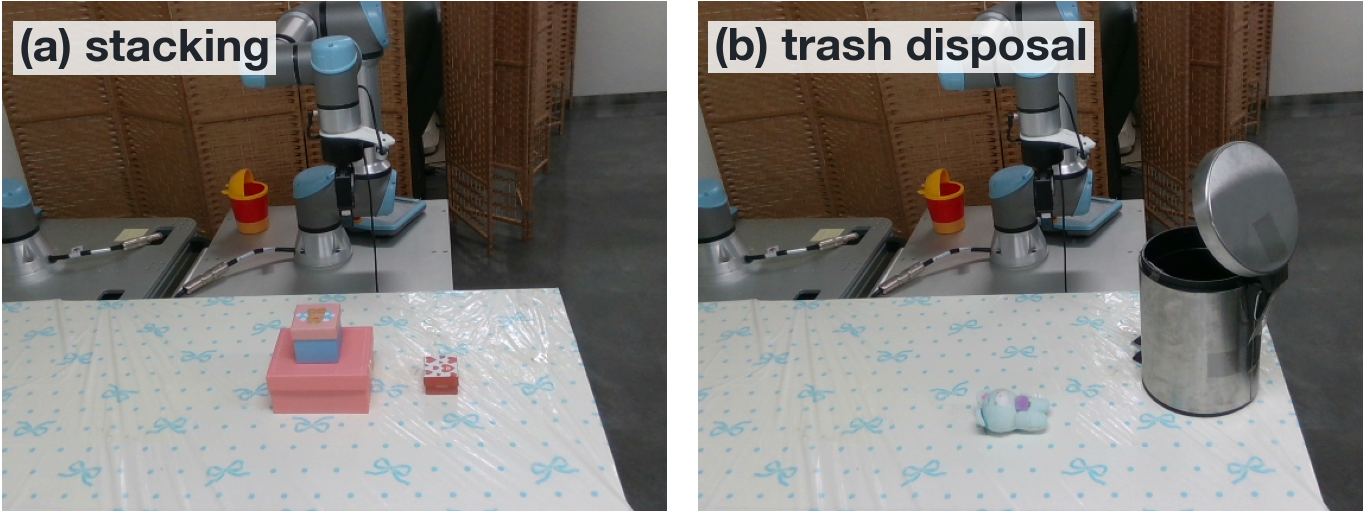}
    \caption{Physical evaluation tasks on the UR5e. (a) Box stacking. (b) Long-range transport into a trash bin.}
    \label{fig:ur5e-tasks}
\end{figure}

We deploy the frozen \pifive{} policy on a UR5e for box stacking and long-range transport into a trash bin (Fig.~\ref{fig:ur5e-tasks}).

We collect 25 matched raw--\method{} pairs per task, yielding 50 pairs and 100 attempts with alternating execution order.
Each raw--\method{} pair shares its policy seed, nominal object and receptacle locations, controller, safety limits, and 300-command budget.

Both conditions use $(H,C)=(10,5)$, the same nominal 3.5-Hz Cartesian controller, and action scales of $0.05$ m and $0.5$ rad per normalized translation and rotation unit.
An independent 125-Hz recorder captures TCP pose and speed, joint velocity, and joint current.
The operator labels stable placement or release as success independently of metric computation.
Each replan logs latency, solver failures, and raw fallbacks.

\begin{table*}[!t]
    \vspace*{4pt}
    \caption{LIBERO performance on the 400-pair \pifive{} holdout.}
    \label{tab:libero-main}
    \centering
    \scriptsize
    \renewcommand{\arraystretch}{0.82}
    \begin{tabular*}{\textwidth}{@{\extracolsep{\fill}}l*{10}{c}@{}}
        \toprule
        \multicolumn{11}{@{}l}{\textit{Executed-EEF jerk reduction from raw (\%).}} \\
        & \multicolumn{2}{c}{Spatial} & \multicolumn{2}{c}{Object} & \multicolumn{2}{c}{Goal} & \multicolumn{2}{c}{LIBERO-10} & \multicolumn{2}{c}{Overall} \\
        \cmidrule(lr){2-3}\cmidrule(lr){4-5}\cmidrule(lr){6-7}\cmidrule(lr){8-9}\cmidrule(l){10-11}
        Method & $\Delta J_{\mathrm{tr}}$ & $\Delta J_{\mathrm{rot}}$ & $\Delta J_{\mathrm{tr}}$ & $\Delta J_{\mathrm{rot}}$ & $\Delta J_{\mathrm{tr}}$ & $\Delta J_{\mathrm{rot}}$ & $\Delta J_{\mathrm{tr}}$ & $\Delta J_{\mathrm{rot}}$ & $\Delta J_{\mathrm{tr}}$ & $\Delta J_{\mathrm{rot}}$ \\
        \midrule
        Raw & -- & -- & -- & -- & -- & -- & -- & -- & -- & -- \\
        RTC & \shortstack{6.19\\[-1pt]{\tiny [4.02, 8.44]}} & \shortstack{7.93\\[-1pt]{\tiny [6.08, 10.08]}} & \shortstack{8.23\\[-1pt]{\tiny [5.59, 10.86]}} & \shortstack{7.44\\[-1pt]{\tiny [4.96, 9.94]}} & \shortstack{4.92\\[-1pt]{\tiny [2.75, 7.22]}} & \shortstack{7.23\\[-1pt]{\tiny [4.78, 9.53]}} & \shortstack{6.11\\[-1pt]{\tiny [0.58, 9.83]}} & \shortstack{7.23\\[-1pt]{\tiny [3.82, 10.06]}} & \shortstack{6.37\\[-1pt]{\tiny [4.73, 7.81]}} & \shortstack{7.46\\[-1pt]{\tiny [6.15, 8.71]}} \\
        SEAM ($\lambda=.05$) & \shortstack{10.20\\[-1pt]{\tiny [8.79, 11.69]}} & \shortstack{13.65\\[-1pt]{\tiny [12.54, 14.87]}} & \shortstack{11.25\\[-1pt]{\tiny [8.32, 14.12]}} & \shortstack{14.59\\[-1pt]{\tiny [12.20, 17.28]}} & \shortstack{8.95\\[-1pt]{\tiny [6.43, 11.75]}} & \shortstack{12.29\\[-1pt]{\tiny [10.17, 14.61]}} & \shortstack{11.78\\[-1pt]{\tiny [7.05, 14.95]}} & \shortstack{14.02\\[-1pt]{\tiny [10.81, 16.34]}} & \shortstack{10.55\\[-1pt]{\tiny [9.04, 11.96]}} & \shortstack{13.64\\[-1pt]{\tiny [12.49, 14.76]}} \\
        SEAM ($\lambda=.10$) & \shortstack{17.95\\[-1pt]{\tiny [15.68, 20.41]}} & \shortstack{21.83\\[-1pt]{\tiny [20.06, 23.62]}} & \shortstack{17.29\\[-1pt]{\tiny [14.20, 20.16]}} & \shortstack{21.73\\[-1pt]{\tiny [18.33, 24.89]}} & \shortstack{14.90\\[-1pt]{\tiny [12.86, 17.53]}} & \shortstack{19.07\\[-1pt]{\tiny [16.74, 21.28]}} & \shortstack{19.44\\[-1pt]{\tiny [14.49, 23.05]}} & \shortstack{22.83\\[-1pt]{\tiny [19.11, 25.83]}} & \shortstack{17.41\\[-1pt]{\tiny [15.79, 18.94]}} & \shortstack{21.38\\[-1pt]{\tiny [19.92, 22.74]}} \\
        SEAM ($\lambda=.20$) & \shortstack{\underline{20.25}\\[-1pt]{\tiny [17.02, 23.74]}} & \shortstack{\underline{24.51}\\[-1pt]{\tiny [22.18, 27.16]}} & \shortstack{\underline{24.94}\\[-1pt]{\tiny [21.83, 27.96]}} & \shortstack{\underline{28.11}\\[-1pt]{\tiny [25.55, 30.40]}} & \shortstack{\underline{18.51}\\[-1pt]{\tiny [15.05, 22.30]}} & \shortstack{\underline{23.62}\\[-1pt]{\tiny [20.39, 26.72]}} & \shortstack{\underline{24.29}\\[-1pt]{\tiny [19.94, 27.89]}} & \shortstack{\underline{27.47}\\[-1pt]{\tiny [23.62, 30.77]}} & \shortstack{\underline{22.05}\\[-1pt]{\tiny [20.27, 23.81]}} & \shortstack{\underline{25.95}\\[-1pt]{\tiny [24.43, 27.40]}} \\
        POTR & \shortstack{10.50\\[-1pt]{\tiny [8.32, 13.15]}} & \shortstack{13.67\\[-1pt]{\tiny [12.18, 15.46]}} & \shortstack{13.27\\[-1pt]{\tiny [11.28, 15.47]}} & \shortstack{13.64\\[-1pt]{\tiny [11.33, 15.89]}} & \shortstack{8.27\\[-1pt]{\tiny [5.11, 11.53]}} & \shortstack{12.02\\[-1pt]{\tiny [9.46, 14.88]}} & \shortstack{11.55\\[-1pt]{\tiny [6.74, 15.15]}} & \shortstack{12.77\\[-1pt]{\tiny [8.83, 15.97]}} & \shortstack{10.92\\[-1pt]{\tiny [9.32, 12.42]}} & \shortstack{13.03\\[-1pt]{\tiny [11.66, 14.31]}} \\
        Overlap ensemble & \shortstack{16.17\\[-1pt]{\tiny [13.67, 18.84]}} & \shortstack{20.36\\[-1pt]{\tiny [18.71, 22.02]}} & \shortstack{21.47\\[-1pt]{\tiny [18.73, 24.12]}} & \shortstack{23.63\\[-1pt]{\tiny [21.04, 26.03]}} & \shortstack{15.63\\[-1pt]{\tiny [13.69, 17.48]}} & \shortstack{19.79\\[-1pt]{\tiny [16.80, 22.49]}} & \shortstack{19.75\\[-1pt]{\tiny [14.90, 23.64]}} & \shortstack{22.67\\[-1pt]{\tiny [19.53, 25.25]}} & \shortstack{18.29\\[-1pt]{\tiny [16.73, 19.76]}} & \shortstack{21.63\\[-1pt]{\tiny [20.31, 22.83]}} \\
        LiPo-QP adapter & \shortstack{14.26\\[-1pt]{\tiny [10.24, 18.48]}} & \shortstack{11.93\\[-1pt]{\tiny [5.97, 17.86]}} & \shortstack{8.99\\[-1pt]{\tiny [5.16, 12.38]}} & \shortstack{8.87\\[-1pt]{\tiny [4.24, 13.25]}} & \shortstack{9.14\\[-1pt]{\tiny [4.84, 13.08]}} & \shortstack{4.50\\[-1pt]{\tiny [-1.93, 10.85]}} & \shortstack{10.86\\[-1pt]{\tiny [6.73, 14.41]}} & \shortstack{7.21\\[-1pt]{\tiny [2.18, 12.34]}} & \shortstack{10.84\\[-1pt]{\tiny [8.81, 12.77]}} & \shortstack{8.17\\[-1pt]{\tiny [5.35, 10.97]}} \\
        \midrule
        \method{} & \shortstack{\textbf{35.32}\\[-1pt]{\tiny [31.65, 38.65]}} & \shortstack{\textbf{46.60}\\[-1pt]{\tiny [42.73, 49.53]}} & \shortstack{\textbf{29.00}\\[-1pt]{\tiny [24.92, 33.07]}} & \shortstack{\textbf{39.47}\\[-1pt]{\tiny [35.28, 43.66]}} & \shortstack{\textbf{31.03}\\[-1pt]{\tiny [26.56, 35.24]}} & \shortstack{\textbf{41.80}\\[-1pt]{\tiny [37.39, 45.66]}} & \shortstack{\textbf{36.44}\\[-1pt]{\tiny [32.47, 40.54]}} & \shortstack{\textbf{45.54}\\[-1pt]{\tiny [41.53, 49.82]}} & \shortstack{\textbf{33.02}\\[-1pt]{\tiny [31.02, 34.97]}} & \shortstack{\textbf{43.42}\\[-1pt]{\tiny [41.38, 45.36]}} \\
        \midrule
        \multicolumn{11}{@{}l}{\textit{Success rate (\%).}} \\
        Method & \multicolumn{2}{c}{Spatial} & \multicolumn{2}{c}{Object} & \multicolumn{2}{c}{Goal} & \multicolumn{2}{c}{LIBERO-10} & \multicolumn{2}{c}{Overall} \\
        \midrule
        Raw & \multicolumn{2}{c}{\shortstack{96.00\\[-1pt]{\tiny [92.00, 100.00]}}} & \multicolumn{2}{c}{\shortstack{97.00\\[-1pt]{\tiny [94.00, 100.00]}}} & \multicolumn{2}{c}{\shortstack{95.00\\[-1pt]{\tiny [90.00, 100.00]}}} & \multicolumn{2}{c}{\shortstack{87.00\\[-1pt]{\tiny [72.00, 98.00]}}} & \multicolumn{2}{c}{\shortstack{93.75\\[-1pt]{\tiny [89.75, 97.00]}}} \\
        RTC & \multicolumn{2}{c}{\shortstack{\underline{97.00} (+1.00)\\[-1pt]{\tiny [94.00, 100.00]}}} & \multicolumn{2}{c}{\shortstack{\underline{99.00} (+2.00)\\[-1pt]{\tiny [97.00, 100.00]}}} & \multicolumn{2}{c}{\shortstack{94.00 (-1.00)\\[-1pt]{\tiny [86.00, 100.00]}}} & \multicolumn{2}{c}{\shortstack{90.00 (+3.00)\\[-1pt]{\tiny [75.00, 100.00]}}} & \multicolumn{2}{c}{\shortstack{95.00 (+1.25)\\[-1pt]{\tiny [90.75, 98.25]}}} \\
        SEAM ($\lambda=.05$) & \multicolumn{2}{c}{\shortstack{94.00 (-2.00)\\[-1pt]{\tiny [89.00, 99.00]}}} & \multicolumn{2}{c}{\shortstack{96.00 (-1.00)\\[-1pt]{\tiny [93.00, 99.00]}}} & \multicolumn{2}{c}{\shortstack{\underline{97.00} (+2.00)\\[-1pt]{\tiny [94.00, 100.00]}}} & \multicolumn{2}{c}{\shortstack{90.00 (+3.00)\\[-1pt]{\tiny [81.00, 97.00]}}} & \multicolumn{2}{c}{\shortstack{94.25 (+0.50)\\[-1pt]{\tiny [91.50, 96.75]}}} \\
        SEAM ($\lambda=.10$) & \multicolumn{2}{c}{\shortstack{\textbf{98.00} (+2.00)\\[-1pt]{\tiny [95.00, 100.00]}}} & \multicolumn{2}{c}{\shortstack{96.00 (-1.00)\\[-1pt]{\tiny [92.00, 100.00]}}} & \multicolumn{2}{c}{\shortstack{\underline{97.00} (+2.00)\\[-1pt]{\tiny [93.00, 100.00]}}} & \multicolumn{2}{c}{\shortstack{90.00 (+3.00)\\[-1pt]{\tiny [83.00, 96.00]}}} & \multicolumn{2}{c}{\shortstack{95.25 (+1.50)\\[-1pt]{\tiny [92.75, 97.50]}}} \\
        SEAM ($\lambda=.20$) & \multicolumn{2}{c}{\shortstack{\textbf{98.00} (+2.00)\\[-1pt]{\tiny [95.00, 100.00]}}} & \multicolumn{2}{c}{\shortstack{\textbf{100.00} (+3.00)\\[-1pt]{\tiny [100.00, 100.00]}}} & \multicolumn{2}{c}{\shortstack{\textbf{98.00} (+3.00)\\[-1pt]{\tiny [94.00, 100.00]}}} & \multicolumn{2}{c}{\shortstack{\underline{93.00} (+6.00)\\[-1pt]{\tiny [87.00, 98.00]}}} & \multicolumn{2}{c}{\shortstack{\textbf{97.25} (+3.50)\\[-1pt]{\tiny [95.25, 99.00]}}} \\
        POTR & \multicolumn{2}{c}{\shortstack{96.00 (+0.00)\\[-1pt]{\tiny [93.00, 99.00]}}} & \multicolumn{2}{c}{\shortstack{\underline{99.00} (+2.00)\\[-1pt]{\tiny [97.00, 100.00]}}} & \multicolumn{2}{c}{\shortstack{95.00 (+0.00)\\[-1pt]{\tiny [89.00, 100.00]}}} & \multicolumn{2}{c}{\shortstack{92.00 (+5.00)\\[-1pt]{\tiny [84.00, 99.00]}}} & \multicolumn{2}{c}{\shortstack{95.50 (+1.75)\\[-1pt]{\tiny [92.75, 97.75]}}} \\
        Overlap ensemble & \multicolumn{2}{c}{\shortstack{96.00 (+0.00)\\[-1pt]{\tiny [92.00, 100.00]}}} & \multicolumn{2}{c}{\shortstack{\underline{99.00} (+2.00)\\[-1pt]{\tiny [97.00, 100.00]}}} & \multicolumn{2}{c}{\shortstack{96.00 (+1.00)\\[-1pt]{\tiny [89.98, 100.00]}}} & \multicolumn{2}{c}{\shortstack{\textbf{94.00} (+7.00)\\[-1pt]{\tiny [88.00, 99.00]}}} & \multicolumn{2}{c}{\shortstack{\underline{96.25} (+2.50)\\[-1pt]{\tiny [93.75, 98.25]}}} \\
        LiPo-QP adapter & \multicolumn{2}{c}{\shortstack{\textbf{98.00} (+2.00)\\[-1pt]{\tiny [94.00, 100.00]}}} & \multicolumn{2}{c}{\shortstack{\underline{99.00} (+2.00)\\[-1pt]{\tiny [97.00, 100.00]}}} & \multicolumn{2}{c}{\shortstack{\underline{97.00} (+2.00)\\[-1pt]{\tiny [93.00, 100.00]}}} & \multicolumn{2}{c}{\shortstack{90.00 (+3.00)\\[-1pt]{\tiny [78.00, 98.00]}}} & \multicolumn{2}{c}{\shortstack{96.00 (+2.25)\\[-1pt]{\tiny [92.50, 98.75]}}} \\
        \midrule
        \method{} & \multicolumn{2}{c}{\shortstack{94.00 (-2.00)\\[-1pt]{\tiny [89.00, 99.00]}}} & \multicolumn{2}{c}{\shortstack{\textbf{100.00} (+3.00)\\[-1pt]{\tiny [100.00, 100.00]}}} & \multicolumn{2}{c}{\shortstack{95.00 (+0.00)\\[-1pt]{\tiny [90.00, 100.00]}}} & \multicolumn{2}{c}{\shortstack{\textbf{94.00} (+7.00)\\[-1pt]{\tiny [88.00, 99.00]}}} & \multicolumn{2}{c}{\shortstack{95.75 (+2.00)\\[-1pt]{\tiny [93.25, 97.75]}}} \\
        \bottomrule
    \end{tabular*}
\end{table*}

\subsection{Metrics and Statistical Analysis}

In simulation, translational jerk is the mean norm of third differences in executed EEF position divided by $\Delta t^3$.
Rotational jerk is the mean norm of second differences in quaternion-derived local angular velocity divided by $\Delta t^2$, with $\Delta t=0.05$ s for both metrics.
Primary jerk endpoints are computed over each complete episode or trial.
The first-50-command sensitivity analysis holds exposure length fixed.

We compute physical translational and rotational jerk from the linear- and angular-velocity triplets in the 125-Hz RTDE \emph{actual\_TCP\_speed} signal.
The analysis spans the first command through 250 ms after the final command.
We resample in controller time, compute second derivatives with a fixed 11-sample third-order Savitzky--Golay filter, trim five samples at each boundary, and average the derivative norms.
TCP travel is the integral of the linear-speed norm.
Joint acceleration and current slew are the mean norms of the first Savitzky--Golay derivatives of RTDE joint velocity and current, using the same resampling, window, and trimming.

For positive endpoints, we report the paired geometric-mean reduction $100[1-\exp(-\overline{\log(J^{\mathrm{raw}}/J^{\mathrm{method}})})]$.
We form 95\% confidence intervals from 10,000 paired bootstrap draws with seed 0.
We resample tasks within suites in simulation and pairs within tasks on the robot, using equal group weights in both analyses.
For task success, we report paired percentage-point changes and use an exact McNemar test.

The primary physical analysis includes measured replanning delays.
For the timing sensitivity analysis, we rescale each endpoint by its mean command interval raised to the derivative order, using three for jerk, two for joint acceleration, and one for current slew.
We apply this standardization to full trials and to strict-interior intervals that exclude the intervals immediately before and after each replan.

\section{Results and Analysis}

\subsection*{Q1. How Does GROOVE Affect EEF Jerk and Task Success?}

\begin{figure}[!t]
    \centering
    \includegraphics[width=\columnwidth]{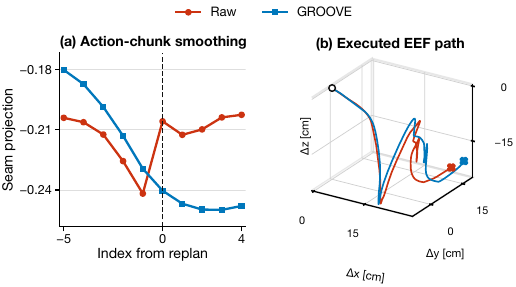}
    \caption{Representative LIBERO pair nearest the median translational-jerk reduction. (a) Raw and delivered \method{} delta-EEF commands projected onto the unit direction of the raw boundary jump. (b) Start-relative executed EEF paths. Circles and crosses denote starts and ends.}
    \label{fig:libero-representative-traces}
\end{figure}

Table~\ref{tab:libero-main} summarizes jerk reduction and task success on the shared holdout.
Parentheses show paired success-rate changes from raw execution, and bold and underlined values mark the best and second-best point estimates, respectively.
Figure~\ref{fig:libero-representative-traces} shows a representative replan and its matched executed path.
\method{} achieves the largest translational and rotational jerk reductions for every LIBERO suite and overall, reducing aggregate EEF jerk by 33.02\% and 43.42\%, respectively.
The strongest SEAM setting, $\lambda=0.20$, reaches 22.05\% and 25.95\% jerk reduction with 97.25\% task success.
The reductions from \method{} are 1.50 and 1.67 times as large under the same policy-query and command budgets.
Task success increases from 93.75\% for raw execution to 95.75\% with \method{}, a paired change of $+2.00$ percentage points with a 95\% confidence interval of $[-0.25,\,4.50]$.
Among the 24 discordant pairs, 16 succeed only with \method{} and 8 succeed only with raw execution, with an exact McNemar $p$-value of 0.152.
Over the first 50 commands, \method{} reduces translational and rotational jerk by \PiCompFixedGrooveTrans{} \PiCompFixedGrooveTransCI{} and \PiCompFixedGrooveRot{} \PiCompFixedGrooveRotCI{}, respectively.
All 400 assignments are completed for each direct comparator, with no solver fallbacks.

On the development grids, the same frozen regulator settings reduce translational and rotational jerk by \PiTransJerk{} and \PiRotJerk{} on \pifive{} and by \GRTransJerk{} and \GRRotJerk{} on \grt{}.
Task success changes by \PiSR{} and \GRSR{}, respectively.

\begin{figure}[!t]
    \centering
    \includegraphics[width=\columnwidth]{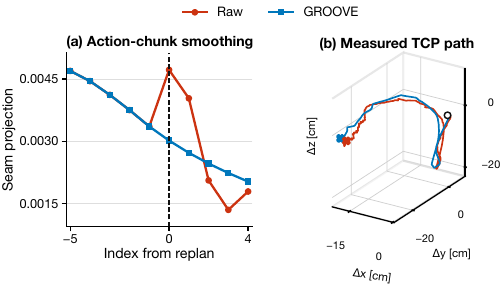}
    \caption{Representative UR5e transport pair nearest the task-median translational-jerk reduction. (a) Raw and delivered \method{} delta-EEF commands projected onto the unit direction of the raw boundary jump. (b) Start-relative measured TCP paths. Circles and crosses denote starts and ends.}
    \label{fig:ur5e-representative-traces}
\end{figure}

\subsection*{Q2. Do the Gains Persist on the Physical Robot?}

On the UR5e, \method{} reduces TCP jerk and actuator-demand variation (Table~\ref{tab:ur5e-effects} and Fig.~\ref{fig:ur5e-representative-traces}).

\begin{table}[!htb]
    \caption{UR5e reductions relative to raw execution.}
    \label{tab:ur5e-effects}
    \centering
    \footnotesize
    \setlength{\tabcolsep}{4pt}
    \begin{tabularx}{\columnwidth}{@{}Xr@{}}
        \toprule
        Metric & Reduction [95\% CI] \\
        \midrule
        Translational TCP jerk & \URTransTCPJerk{} \URTransTCPJerkCI{} \\
        Rotational TCP jerk & \URRotTCPJerk{} \URRotTCPJerkCI{} \\
        TCP travel & \URPath{} \URPathCI{} \\
        Joint acceleration & \URJointAccelerationChange{} \URJointAccelerationChangeCI{} \\
        Joint-current slew & \URCurrentSlew{} \URCurrentSlewCI{} \\
        \bottomrule
    \end{tabularx}
\end{table}

Across 9-, 11-, and 15-sample Savitzky--Golay windows, the observed reductions are \URFilterSensitivityTrans{} in translational TCP jerk, \URFilterSensitivityRot{} in rotational TCP jerk, and \URFilterSensitivityCurrent{} in joint-current slew.

Task success changes from \URRawSR{} to \URMethodSR{}, yielding \URSR{} with a 95\% confidence interval of \URSRCI{} and an exact McNemar $p$-value of \URMcNemarP{}.
Stacking improves by \URStackSRChange{}, while long-range transport remains unchanged.

\textbf{Timing sensitivity.}
Across the full trials, the mean command interval increases from \URRawCommandIntervalMean{} to \URMethodCommandIntervalMean{}, a difference of \URCommandIntervalIncrease{}.
The full-trial cadence-standardized estimates are \URFullPaceTrans{} \URFullPaceTransCI{} for translational jerk and \URFullPaceRot{} \URFullPaceRotCI{} for rotational jerk, with both confidence intervals including zero.
The primary effects in Table~\ref{tab:ur5e-effects} include the deployed system's timing. We examine motion between replans separately.

A post-hoc strict-interior analysis excludes the intervals immediately before and after each replan. In the remaining intervals, the command-interval difference is \URInteriorCommandIntervalIncrease{}.
After cadence standardization, translational and rotational jerk remain lower by \URInteriorTrans{} \URInteriorTransCI{} and \URInteriorRot{} \URInteriorRotCI{}, respectively, while joint-current slew decreases by \URInteriorCurrentSlew{} \URInteriorCurrentSlewCI{}.
Uniform time dilation at the observed interval ratio predicts only \URInteriorPaceOnly{} for jerk and \URInteriorCurrentSlewPaceOnly{} for current slew, below the measured reductions.

\subsection*{Q3. Ablation Studies and Motion Fidelity}

\textbf{Candidate-set composition.}
Table~\ref{tab:candidate-set-ablation} reports jerk reductions relative to the follow-up raw rerun, which achieves 95.50\% success.
Relative to the cube alone, adding three coordinate axes or four body diagonals reduces translational jerk by 2.72\% and 3.73\%, respectively.
The full set reduces it by $5.33\%~[3.25,\,7.24]$.
Relative to the full set, the four-diagonal subset has $1.69\%~[-0.39,\,3.73]$ higher translational jerk and a success difference of $+0.25~[-1.25,\,1.75]$ percentage points.
The full set's incremental benefit over four diagonals remains uncertain in LIBERO.
Selection changes translation only, so rotational differences arise through closed-loop replanning.

\textbf{Deviation allowance.}
In the 400-triplet rollout, the full set reduces translational jerk from cube-only execution by $1.60\%~[0.08,\,3.22]$ at $\rho_B=1$ and $6.33\%~[4.27,\,8.41]$ at $\rho_B=1.5$.
Success is 96.0\% and 95.5\%, respectively, versus 95.5\% for cube-only execution.

\textbf{Matched-deviation replay.}
Across 11,356 matched replans (87.5\%) from all 400 episodes, the full set lowers command-space jerk by $3.28\%~[2.92,\,3.65]$ relative to cubes with matched realized prefix deviation.

\begin{table}[H]
    \caption{Candidate-set ablation on the \pifive{} LIBERO follow-up.}
    \label{tab:candidate-set-ablation}
    \centering
    \scriptsize
    \setlength{\tabcolsep}{2pt}
    \begin{tabularx}{\columnwidth}{@{}Xcccc@{}}
        \toprule
        Setting & \shortstack{No. of\\candidates} & \shortstack{SR\\{[\%]}} & \shortstack{$\Delta J_{\mathrm{tr}}$ [\%]\\[-1pt]{\tiny 95\% CI}} & \shortstack{$\Delta J_{\mathrm{rot}}$ [\%]\\[-1pt]{\tiny 95\% CI}} \\
        \midrule
        Cube ($k=0$) & 1 & 95.50 & \shortstack{$29.28$\\{\tiny $[27.03,\,31.39]$}} & \shortstack{$42.02$\\{\tiny $[39.68,\,44.28]$}} \\
        Cube + 3 axes & 4 & 94.50 & \shortstack{$31.21$\\{\tiny $[29.04,\,33.39]$}} & \shortstack{$42.47$\\{\tiny $[40.34,\,44.54]$}} \\
        Cube + 4 body diagonals & 5 & 96.25 & \shortstack{$31.92$\\{\tiny $[29.67,\,34.19]$}} & \shortstack{$42.91$\\{\tiny $[40.63,\,45.18]$}} \\
        Full set & 14 & 96.00 & \shortstack{$33.05$\\{\tiny $[31.06,\,35.03]$}} & \shortstack{$43.74$\\{\tiny $[41.55,\,45.93]$}} \\
        \bottomrule
    \end{tabularx}
\end{table}

\textbf{Candidate selection.}
On the \pifive{} development grid, the full set yields lower translational and rotational jerk than one random direction, and protected selection yields lower jerk than random eligible selection (Table~\ref{tab:mechanism-controls}, first two rows).

\begin{table}[H]
    \caption{Selection (development grid) and fidelity (fixed-50 holdout) ablations with \pifive{}.}
    \label{tab:mechanism-controls}
    \centering
    \scriptsize
    \setlength{\tabcolsep}{2pt}
    \begin{tabularx}{\columnwidth}{@{}Xccc@{}}
        \toprule
        Contrast & \shortstack{$\Delta J_{\mathrm{tr}}$ [\%]\\[-1pt]{\tiny 95\% CI}} & \shortstack{$\Delta J_{\mathrm{rot}}$ [\%]\\[-1pt]{\tiny 95\% CI}} & \shortstack{$\Delta\mathrm{SR}$ [pp]\\[-1pt]{\tiny 95\% CI}} \\
        \midrule
        Full set vs. random direction & \shortstack{$+7.83$\\{\tiny $[6.95,\,8.63]$}} & \shortstack{$+5.85$\\{\tiny $[4.64,\,6.98]$}} & \shortstack{$-0.75$\\{\tiny $[-1.88,\,0.38]$}} \\
        Protected selector vs. random eligible & \shortstack{$+3.29$\\{\tiny $[2.06,\,4.53]$}} & \shortstack{$+2.66$\\{\tiny $[1.47,\,3.89]$}} & \shortstack{$+0.25$\\{\tiny $[-1.00,\,1.50]$}} \\
        Every-prefix vs. per-command & \shortstack{$+10.28$\\{\tiny $[8.29,\,12.25]$}} & \shortstack{$+2.75$\\{\tiny $[-1.27,\,6.46]$}} & \shortstack{$+3.25$\\{\tiny $[0.00,\,7.00]$}} \\
        \bottomrule
    \end{tabularx}
\end{table}

\textbf{History and fidelity constraints.}
On the separate 400-pair fixed-50-command holdout, the cube regulator reduces translational jerk by $38.52\%~[35.58,\,41.59]$ relative to raw execution and by $16.14\%~[12.93,\,19.37]$ and $16.53\%~[13.40,\,19.70]$ relative to budget-matched EMA and One Euro filters.
The every-prefix guard reduces translational jerk by 10.28\% relative to the per-command guard.
The rotational reduction is 2.75\%, with a 95\% interval spanning zero (Table~\ref{tab:mechanism-controls}, last row).
A comparison with the no-history control shows a descriptive $35.20\%~[31.53,\,39.05]$ difference in favor of history conditioning.
The endpoint-only guard lowers jerk by 25.14\% relative to the every-prefix regulator but increases mean realized prefix deviation per replan by 90.7\% and lowers success by 4.75 percentage points, illustrating the tradeoff between jerk reduction and motion fidelity.

\subsection*{Q4. What Is the Runtime Overhead?}

With the shared-rotation implementation (Sec.~\ref{sec:groove-method}), mean delays from policy output to command readiness are \URRawOnlinePostPolicyMean{} for raw execution and \UROnlinePostPolicyMean{} for \method{}.
The 95th percentile of this delay for \method{} is \UROnlinePostPolicyPninetyFive{}.
These delays include regulation, diagnostics, logging, and runtime contention.
Exact-input replay of all \URReplayReplans{} replans on the same A100 host reproduces every action.
The regulator core averages \URReplayMean{}, with a 95th percentile of \URReplayPninetyFive{} (\URReplayPninetyFivePeriodFraction{} of the nominal command period).
No solver failure or raw fallback occurs during deployment.

\section{Conclusion}

\method{} regulates frozen VLA chunks without retraining or additional inference, reducing boundary and within-chunk jerk under every-prefix translation and local-coordinate rotation bounds.
It yields the largest jerk reductions among the evaluated methods on the shared LIBERO holdout and lowers measured TCP jerk and current slew across 50 matched UR5e pairs.
Both evaluations show higher aggregate task-success point estimates.
We will make the code and supporting materials publicly available upon publication.

\section*{Acknowledgments}

This work was supported by Samsung Research Funding \& Incubation Center of Samsung Electronics under Project Number SRFC-IT2402-05 and by the Institute of Information \& Communications Technology Planning \& Evaluation (IITP) grant funded by the Korea government (MSIT) (IITP-202027-RS-2026-25616370, AI Star Fellowship Support Program).

\bibliographystyle{IEEEtran}
\bibliography{references}

@IEEEtranBSTCTL{IEEEexample:BSTcontrol,
  CTLuse_forced_etal       = {yes},
  CTLmax_names_forced_etal = {6},
  CTLnames_show_etal       = {1}
}

@inproceedings{act,
  author    = {Tony Z. Zhao and Vikash Kumar and Sergey Levine and Chelsea Finn},
  title     = {Learning Fine-Grained Bimanual Manipulation with Low-Cost Hardware},
  booktitle = {Proceedings of Robotics: Science and Systems},
  year      = {2023},
  address   = {Daegu, Republic of Korea},
  month     = jul,
  doi       = {10.15607/RSS.2023.XIX.016}
}

@inproceedings{bid,
  author    = {Yuejiang Liu and Jubayer Ibn Hamid and Annie Xie and Yoonho Lee and Maximilian Du and Chelsea Finn},
  title     = {Bidirectional Decoding: Improving Action Chunking via Guided Test-Time Sampling},
  booktitle = {International Conference on Learning Representations},
  year      = {2025},
  url       = {https://openreview.net/forum?id=qZmn2hkuzw}
}

@inproceedings{rtc,
  author    = {Kevin Black and Manuel Y. Galliker and Sergey Levine},
  title     = {Real-Time Execution of Action Chunking Flow Policies},
  booktitle = {Advances in Neural Information Processing Systems},
  volume    = {38},
  pages     = {37596--37620},
  year      = {2025},
  doi       = {10.52202/085713-1122}
}

@inproceedings{paint,
  author        = {Trong-Bao Ho and Quang-Tan Nguyen and Thien-Loc Ha and Gia-Binh Nguyen and Viet-Thanh Nguyen and Long Dinh and Minh N. Vu and Duy M. H. Nguyen and An Thai Le and Ngo Anh Vien},
  title         = {Start Right, Arrive Right: Asynchronous Execution via Initial Noise Selection},
  booktitle     = {Conference on Robot Learning},
  year          = {2026},
  note          = {Accepted},
  url           = {https://arxiv.org/abs/2606.19774}
}

@inproceedings{acg,
  author    = {Minho Park and Kinam Kim and Junha Hyung and Hyojin Jang and Hoiyeong Jin and Jooyeol Yun and Hojoon Lee and Jaegul Choo},
  title     = {{ACG}: Action Coherence Guidance for Flow-based Vision-Language-Action models},
  booktitle = {2026 IEEE International Conference on Robotics and Automation (ICRA)},
  year      = {2026},
  url       = {https://davian-robotics.github.io/ACG/}
}

@misc{pace,
  author        = {Junnan Nie and Jiayi Li and Jiachen Zhang and Junyi Lao and Chenghao Liu and Tianle Zhang and Liang Lin and Songfang Huang},
  title         = {{PACE}: Phase-Aware Chunk Execution for Robot Policies with Action Chunking},
  year          = {2026},
  howpublished  = {arXiv preprint arXiv:2606.00537},
  eprint        = {2606.00537},
  archiveprefix = {arXiv},
  primaryclass  = {cs.RO}
}

@inproceedings{aac,
  author    = {Yuanchang Liang and Xiaobo Wang and Kai Wang and Shuo Wang and Xiaojiang Peng and Haoyu Chen and David Kim Huat Chua and Prahlad Vadakkepat},
  title     = {Adaptive Action Chunking at Inference-time for Vision-Language-Action Models},
  booktitle = {Proceedings of the IEEE/CVF Conference on Computer Vision and Pattern Recognition},
  pages     = {20802--20811},
  year      = {2026},
  month     = jun,
  url       = {https://arxiv.org/abs/2604.04161}
}

@misc{a2c2,
  author        = {Kohei Sendai and Maxime Alvarez and Tatsuya Matsushima and Yutaka Matsuo and Yusuke Iwasawa},
  title         = {Leave No Observation Behind: Real-time Correction for {VLA} Action Chunks},
  year          = {2025},
  howpublished  = {arXiv preprint arXiv:2509.23224},
  eprint        = {2509.23224},
  archiveprefix = {arXiv},
  primaryclass  = {cs.RO}
}

@article{lipo,
  author  = {Dongwoo Son and Suhan Park},
  title   = {{LiPo}: A Lightweight Post-optimization Framework for Smoothing Action Chunks Generated by Learned Policies},
  journal = {International Journal of Control, Automation and Systems},
  volume  = {23},
  number  = {11},
  pages   = {3284--3292},
  year    = {2025},
  doi     = {10.1007/s12555-025-0537-0}
}

@inproceedings{openvla,
  author    = {Moo Jin Kim and Karl Pertsch and Siddharth Karamcheti and Ted Xiao and Ashwin Balakrishna and Suraj Nair and Rafael Rafailov and Ethan P. Foster and Pannag R. Sanketi and Quan Vuong and Thomas Kollar and Benjamin Burchfiel and Russ Tedrake and Dorsa Sadigh and Sergey Levine and Percy Liang and Chelsea Finn},
  title     = {{OpenVLA}: An Open-Source Vision-Language-Action Model},
  booktitle = {Proceedings of the 8th Conference on Robot Learning},
  series    = {Proceedings of Machine Learning Research},
  volume    = {270},
  pages     = {2679--2713},
  year      = {2025},
  editor    = {Pulkit Agrawal and Oliver Kroemer and Wolfram Burgard},
  publisher = {PMLR},
  url       = {https://proceedings.mlr.press/v270/kim25c.html}
}

@inproceedings{pi05,
  author    = {Kevin Black and Noah Brown and James Darpinian and Karan Dhabalia and Danny Driess and Adnan Esmail and Michael Robert Equi and Chelsea Finn and Niccolo Fusai and Manuel Y. Galliker and Dibya Ghosh and Lachy Groom and Karol Hausman and brian ichter and Szymon Jakubczak and Tim Jones and Liyiming Ke and Devin LeBlanc and Sergey Levine and Adrian Li-Bell and Mohith Mothukuri and Suraj Nair and Karl Pertsch and Allen Z. Ren and Lucy Xiaoyang Shi and Laura Smith and Jost Tobias Springenberg and Kyle Stachowicz and James Tanner and Quan Vuong and Homer Walke and Anna Walling and Haohuan Wang and Lili Yu and Ury Zhilinsky},
  title     = {{$\pi_{0.5}$}: a Vision-Language-Action Model with Open-World Generalization},
  booktitle = {Proceedings of The 9th Conference on Robot Learning},
  pages     = {17--40},
  year      = {2025},
  editor    = {Joseph Lim and Shuran Song and Hae-Won Park},
  volume    = {305},
  series    = {Proceedings of Machine Learning Research},
  month     = {27--30 September},
  publisher = {PMLR},
  url       = {https://proceedings.mlr.press/v305/black25a.html}
}

@misc{gr00t,
  author        = {{NVIDIA} and Johan Bjorck and Fernando Casta{\~n}eda and Nikita Cherniadev and Xingye Da and Runyu Ding and Linxi {``Jim''} Fan and Yu Fang and Dieter Fox and Fengyuan Hu and Spencer Huang and Joel Jang and Zhenyu Jiang and Jan Kautz and Kaushil Kundalia and Lawrence Lao and Zhiqi Li and Zongyu Lin and Kevin Lin and Guilin Liu and Edith Llontop and Loic Magne and Ajay Mandlekar and Avnish Narayan and Soroush Nasiriany and Scott Reed and You Liang Tan and Guanzhi Wang and Zu Wang and Jing Wang and Qi Wang and Jiannan Xiang and Yuqi Xie and Yinzhen Xu and Zhenjia Xu and Seonghyeon Ye and Zhiding Yu and Ao Zhang and Hao Zhang and Yizhou Zhao and Ruijie Zheng and Yuke Zhu},
  title         = {{GR00T N1}: An Open Foundation Model for Generalist Humanoid Robots},
  year          = {2025},
  howpublished  = {arXiv preprint arXiv:2503.14734},
  eprint        = {2503.14734},
  archiveprefix = {arXiv},
  primaryclass  = {cs.RO}
}

@misc{gr00tn17libero,
  author       = {{NVIDIA}},
  title        = {{GR00T-N1.7-LIBERO}},
  year         = {2026},
  howpublished = {Hugging Face model card and checkpoint},
  url          = {https://huggingface.co/nvidia/GR00T-N1.7-LIBERO},
  note         = {Accessed August 29, 2026}
}

@inproceedings{libero,
  author    = {Bo Liu and Yifeng Zhu and Chongkai Gao and Yihao Feng and Qiang Liu and Yuke Zhu and Peter Stone},
  title     = {{LIBERO}: Benchmarking Knowledge Transfer for Lifelong Robot Learning},
  booktitle = {Advances in Neural Information Processing Systems},
  volume    = {36},
  pages     = {44776--44791},
  year      = {2023},
  doi       = {10.52202/075280-1939}
}

@misc{seam,
  author        = {Dijia Zhan and Xuemiao Xu and Jinyi Li and Jie Tang},
  title         = {{SEAM}: Smooth Execution of Action-Chunked Motion for Vision-Language-Action Policies},
  year          = {2026},
  howpublished  = {arXiv preprint arXiv:2607.04609},
  eprint        = {2607.04609},
  archiveprefix = {arXiv},
  primaryclass  = {cs.RO}
}

@misc{chunkflow,
  author        = {Zhao Yang and Yinan Shi and Mingyuan Yao and Wenyao Xue and Yawei Jueluo and Longjun Liu},
  title         = {{ChunkFlow}: Towards Continuity-Consistent Chunked Policy Learning},
  year          = {2026},
  howpublished  = {arXiv preprint arXiv:2607.12992},
  eprint        = {2607.12992},
  archiveprefix = {arXiv},
  primaryclass  = {cs.RO}
}

@misc{potr,
  author        = {Kai Fang and Hailong Pei and Xuemin Chi},
  title         = {Smoother Action Chunking Flow Policy via Prior-Corrected Orthogonal Trust-Region Guidance},
  year          = {2026},
  howpublished  = {arXiv preprint arXiv:2605.24433},
  eprint        = {2605.24433},
  archiveprefix = {arXiv},
  primaryclass  = {cs.RO}
}

@misc{embodiedefficiency,
  author        = {Zhuofan Li and Hongkun Yang and Zhenyang Chen and Yangxuan Chen and Yingyan {(Celine)} Lin and Chaojian Li},
  title         = {From Inference Efficiency to Embodied Efficiency: Revisiting Efficiency Metrics for Vision-Language-Action Models},
  year          = {2026},
  howpublished  = {arXiv preprint arXiv:2603.19131},
  eprint        = {2603.19131},
  archiveprefix = {arXiv},
  primaryclass  = {cs.LG}
}

@misc{smoothvla,
  author        = {Jiashun Li and Xiaoyu Shi and Hong Xie and Mingsheng Shang and Yun Lu},
  title         = {{SmoothVLA}: Aligning Vision-Language-Action Models with Physical Constraints via Intrinsic Smoothness Optimization},
  year          = {2026},
  howpublished  = {arXiv preprint arXiv:2603.13925},
  eprint        = {2603.13925},
  archiveprefix = {arXiv},
  primaryclass  = {cs.RO}
}

@inproceedings{legato,
  author    = {Yufeng Liu and Hang Yu and Juntu Zhao and Bocheng Li and Di Zhang and Mingzhu Li and Wenxuan Wu and Yingdong Hu and Junyuan Xie and Junliang Guo and Dequan Wang and Yang Gao},
  title     = {Learning Native Continuation for Action Chunking Flow Policies},
  booktitle = {Proceedings of Robotics: Science and Systems},
  year      = {2026},
  doi       = {10.15607/RSS.2026.XXII.058},
  url       = {https://roboticsproceedings.org/rss22/p058.html}
}

@inproceedings{rtr,
  author    = {Kunyun Wang and Yuhang Zheng and Yupeng Zheng and Jieru Zhao and Wenchao Ding},
  title     = {Learning High-Frequency Continuous Action Chunks in Latent Space},
  booktitle = {Proceedings of the 43rd International Conference on Machine Learning},
  year      = {2026},
  url       = {https://arxiv.org/abs/2605.24931}
}

@inproceedings{oneeuro,
  author    = {G{\'e}ry Casiez and Nicolas Roussel and Daniel Vogel},
  title     = {{1\,\texteuro{} Filter}: A Simple Speed-based Low-pass Filter for Noisy Input in Interactive Systems},
  booktitle = {Proceedings of the SIGCHI Conference on Human Factors in Computing Systems},
  pages     = {2527--2530},
  year      = {2012},
  address   = {New York, NY, USA},
  month     = may,
  publisher = {ACM},
  doi       = {10.1145/2207676.2208639}
}

@article{flashhogan,
  author  = {Tamar Flash and Neville Hogan},
  title   = {The coordination of arm movements: an experimentally confirmed mathematical model},
  journal = {The Journal of Neuroscience},
  volume  = {5},
  number  = {7},
  pages   = {1688--1703},
  year    = {1985},
  doi     = {10.1523/JNEUROSCI.05-07-01688.1985}
}

@article{bearee2014dampedjerk,
  author  = {Richard B{\'e}ar{\'e}e},
  title   = {New Damped-Jerk Trajectory for Vibration Reduction},
  journal = {Control Engineering Practice},
  volume  = {28},
  pages   = {112--120},
  year    = {2014},
  doi     = {10.1016/j.conengprac.2014.03.010}
}

@inproceedings{lee2024jerkconstrained,
  author    = {Jee-Eun Lee and Andrew Bylard and Robert Sun and Luis Sentis},
  title     = {On the Performance of Jerk-Constrained Time-Optimal Trajectory Planning for Industrial Manipulators},
  booktitle = {2024 IEEE International Conference on Robotics and Automation (ICRA)},
  pages     = {9772--9778},
  year      = {2024},
  doi       = {10.1109/ICRA57147.2024.10610437}
}

@inproceedings{ruckig,
  author    = {Lars Berscheid and Torsten Kr{\"o}ger},
  title     = {Jerk-limited Real-time Trajectory Generation with Arbitrary Target States},
  booktitle = {Proceedings of Robotics: Science and Systems},
  year      = {2021},
  address   = {Virtual},
  month     = jul,
  doi       = {10.15607/RSS.2021.XVII.015}
}

@article{osqp,
  author  = {Bartolomeo Stellato and Goran Banjac and Paul Goulart and Alberto Bemporad and Stephen Boyd},
  title   = {{OSQP}: an operator splitting solver for quadratic programs},
  journal = {Mathematical Programming Computation},
  volume  = {12},
  number  = {4},
  pages   = {637--672},
  year    = {2020},
  doi     = {10.1007/s12532-020-00179-2}
}

@inproceedings{realtimevlav2,
  author        = {Chen Yang and Yucheng Hu and Yunchao Ma and Yunhuan Yang and Jing Tan and Haoqiang Fan},
  title         = {{Realtime-VLA V2}: Learning to Run {VLAs} Fast, Smooth, and Accurate},
  booktitle     = {Proceedings of the IEEE/CVF Conference on Computer Vision and Pattern Recognition (CVPR) Workshops},
  pages         = {4501--4509},
  year          = {2026},
  month         = jun,
  url           = {https://arxiv.org/abs/2603.26360}
}

\end{document}